# Helix 1.0: An Open-Source Framework for Reproducible and Interpretable Machine Learning on Tabular Scientific Data

Eduardo Aguilar-Bejarano[1], Daniel Lea[1], Karthikeyan Sivakumar[1], Jimiama M. Mase[1], Reza Omidvar[1], Ruizhe Li[2], Troy Kettle[2], James Mitchell-White[1,3], Morgan R Alexander[4], David A Winkler[5,6], Grazziela Figueredo[3]

1. Digital Research Service, University of Nottingham
2. School of Computer Science, University of Nottingham
3. Centre for Health Informatics, School of Medicine, University of Nottingham
4. School of Pharmacy, University of Nottingham
5. Department of Biochemistry and Chemistry, La Trobe Institute for Molecular Sciences, La Trobe University, Melbourne, Victoria 3086, Australia
6. Monash Institute of Pharmaceutical Sciences, Monash University, Parkville, Victoria 3052, Australia

**Abstract**

Helix is an open-source, extensible, Python-based software framework to facilitate reproducible and interpretable machine learning workflows for tabular data. It addresses the growing need for transparent experimental data analytics provenance, ensuring that the entire analytical process—including decisions around data transformation and methodological choices—is documented, accessible, reproducible, and comprehensible to relevant stakeholders. The platform comprises modules for standardised data preprocessing, visualisation, machine learning model training, evaluation, interpretation, results inspection, and model prediction for unseen data. To further empower researchers without formal training in data science to derive meaningful and actionable insights, Helix features a user-friendly interface that enables the design of computational experiments, inspection of outcomes, including a novel interpretation approach to machine learning decisions using linguistic terms — all within an integrated environment. Released under the MIT licence, Helix is accessible via GitHub and PyPI, supporting community-driven development and promoting adherence to the FAIR principles.

**Highlights**

- Helix is an open source, modular, extensible end-to-end architecture for tabular data analysis and interpretable machine learning
- Emphasis on scientific transparency and reproducibility
- Balance usability, flexibility and methodological rigour, lowering the entry barrier for domain scientists
- Integrated interpretability tools that produce rules in natural language
- Focus on provenance-aware experimentation

# 1. Introduction

The massive increase in data in scientific research requires the development and application of robust tools for data analysis and machine learning (ML) that are findable, accessible, interoperable, reusable (FAIR) and interpretable. In domains, such as biomaterials science, engineering, chemistry, healthcare and biosciences, data-driven discovery typically requires interdisciplinary teams. These teams collaborate to implement unbiased data pre-processing strategies, select appropriate modelling techniques, and interpret model outputs to accelerate and inform research outcomes and support rational design and decision-making. This process is often iterative, with experts providing feedback over long periods of time to refine models and optimise the methodology adopted. In cases where initial analysis identifies issues with the data, such as outliers, unbalanced data classes, or experimental measurement uncertainty, another round of data collection and pre-processing might be necessary. That means that data for the same problem are likely to be analysed multiple times using different dataset versions and methodological pipelines.

For interdisciplinary co-development of analytics, there is also a need for tools that allow domain experts to focus on interpreting and using analysis results, rather than developing code. The widespread use of ML and the overwhelming availability of thousands of community-driven open-source packages in Python and R increases the barrier for interoperable and reusable data analysis methodologies. To facilitate accurate analytics, transparency, and modelling results comparison, there is a strong need for easy-to-use tools that automatically track data, all methodological choices, performance metrics, and corresponding results. Recording provenance facilitates rigorous reporting and fosters confidence in the reliability and replication of outcomes.

Focussing on the FAIR use of ML, Samuel et al. conducted a survey of experts and identified challenges around general ML reproducibility.[1] Those relevant to provenance include availability of the source code used, information on data used for training, testing, and evaluation, lack of reference implementation, insufficient model parameter description, missing information on software requirements (packages used, package versions, dependencies for instance), changes in the code not reflected in the final publication, missing information on modelling methods used and lack of documentation on data pre-processing.

While numerous tools exist to support FAIR-aligned analytics and ML workflows for tabular data, they often lack the flexibility or transparency required for rigorous end-to-end team-driven data analysis. Instead, they mostly focus on FAIR data and ML deployment and are often research domain-specific. Hopsworks[2], for instance, is an open-source platform that integrates big data versioning, metadata tracking, and feature management, facilitating the deployment of ML pipelines with a focus on FAIR principles. It emphasises data provenance and reproducibility, ensuring that datasets and models are well-documented and accessible. It presents however a higher entry barrier for non-technical users. Its open-source version also limits, for example, the number of models that can be deployed. ProvBook[1] is a proof-of-concept tool to enhance reproducibility of ML experiments using Jupyter notebooks. Although it captures the provenance of data and computational steps taken, it relies on users understanding Python. It aims to demonstrate the importance of ML modelling provenance, rather than providing an end-to-end tool for analysis. AIF360[3] and FAT Forensics[4] target fairness auditing and interpretability of MLs. They, however, do not address other aspects of the analytics pipeline, such as data pre-processing. Dalex[5] is a Python package that provides tools for visualising and explaining ML models. It supports the interpretability of complex models, making them more transparent, facilitating the understanding and reuse of ML models. SIMON, on the other hand, is an analytics tool that addresses the need for end-to-end analysis, and includes data pre-processing, statistical tests and ML modelling for biomedical data.[6] It uses a graphical interface that facilitates running the analysis, focussing on supporting results interpretation. It was developed using R but has a few limitations regarding customisation and post-training model interpretation.

Understanding the rationale behind machine learning (ML) model decisions is critical, particularly in systems where verification, regulatory compliance, ethics, and trust are very important. Post hoc techniques such as feature importance (FI) analysis are widely used to interpret ML outputs by quantifying the contribution of each input variable to the model's prediction. Accurate FI estimation can support causal inference, domain expert validation, and identification of biases, thereby enhancing model transparency and fairness. However, the diversity of ML algorithms and FI techniques introduces variability and uncertainty in interpretations. Differences in model structure, learning strategies, and the local/nonlinear nature of data can lead to inconsistent FI results. So, there is the also a need for tools that are easy to use and can address this uncertainty in model interpretation.

## 2. Results

We propose that Helix is a suitable framework for addressing the above challenges. It offers an open-source solution that streamlines the experimental analysis process from data input to model interpretation and deployment, while keeping a record of the entire analytics decision provenance. Although originally developed for quantitative structure–property relationship (QSPR) modelling in biomaterials discovery, it is broadly applicable to any tabular classification or regression tasks in many domains. Helix's functional requirements were driven by the need for a user-friendly, flexible, extensible platform that allows experimentation with different analytics pipelines.  It aims to be accessible for experimental specialists with little or no formal data analytics training.

Helix was implemented using research software engineering best practices. It adopts an object-oriented architecture to facilitate modularity, extensibility, and maintainability.  It allows for integration of ML models, pre-processing and interpretation techniques, facilitating adaptability for future needs. The general analytics workflow of Helix is shown in Figure 1.

Once an experiment is created and data is uploaded, the user can apply several pre-processing and visualisation approaches to the data (Stage 1 in Figure 1). Helix provides tools for data normalisation, transformation, and feature selection. It includes functionalities for generating plots, statistics and normality tests, aiding exploratory data analysis. Raw or pre-processed data are then used for machine learning modelling (Stage 2). Users can train and evaluate multiple ML models with options for data training/testing splits and hyperparameter tuning. Helix implements ensemble feature importance methods introduced in[7] (Stage 3) and[8] (Stage 4), combining outputs from various models to identify key predictors and linguistic rules that describe the synergistic importance of independent variables within context.

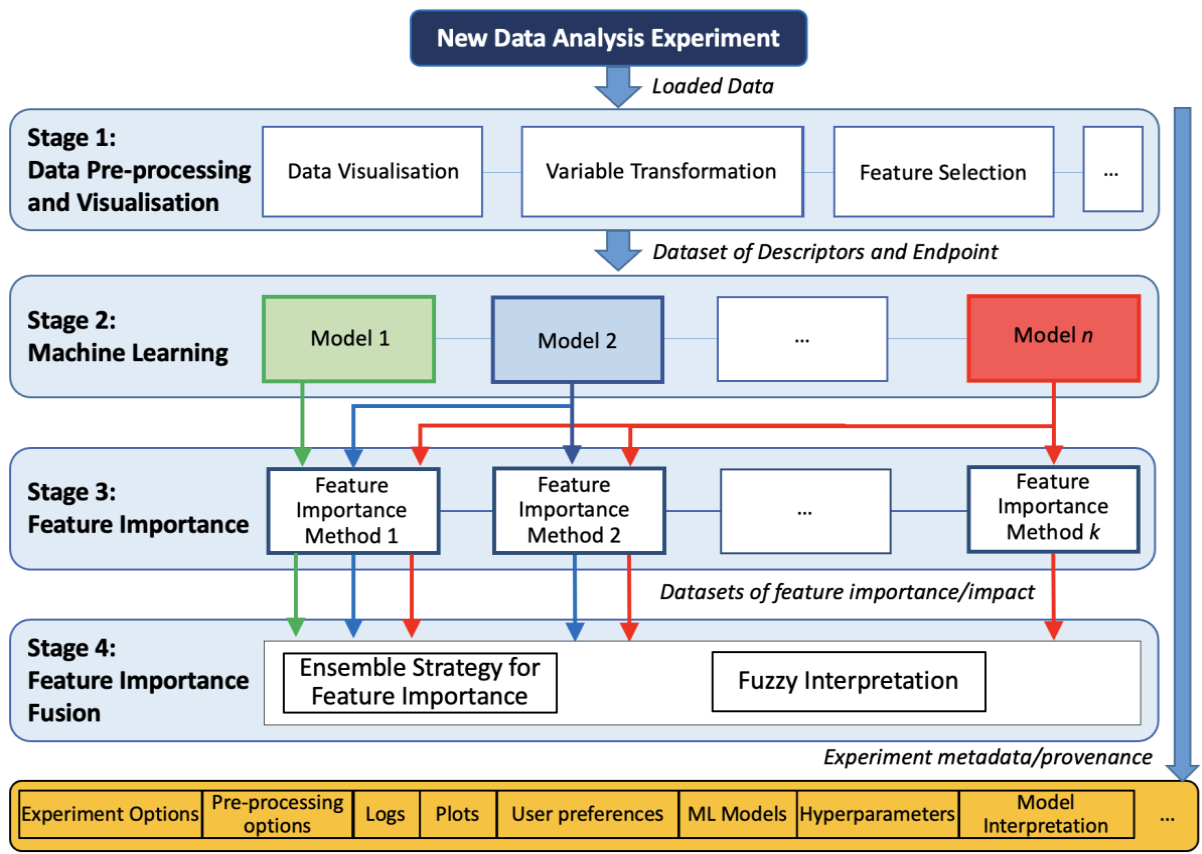

*Figure 1- Helix general analytics workflow and provenance recording*

## 2.1 Object Oriented Architecture

Helix data operations are encapsulated in classes that manage structured data manipulation and transformation. The ML implementation has a *Learner* base class, which abstracts essential operations for model training, including support for cross-validation (holdout and K-fold), bootstrap sampling, and model evaluation using standard metrics. This class enables the addition of user-defined models via inheritance and serves as a foundation for hyperparameter optimisation. Both classes conform to a unified interface, enabling integration of classification and regression approaches. Interpretability is managed by the *FeatureImportanceEstimator* class, which supports both global and local instance-based feature importance analyses. This class employs ensemble methods and allows for the addition of alternative interpretability approaches. There is a services layer, which comprises a suite of functional modules for statistical testing, metric computation, model instantiation, and data preprocessing. These services are decoupled from the domain logic and are injected as dependencies, promoting loose coupling and testability. User interface components and modelling options are similarly modularised, with user input forms logic and visualisation capabilities segregated into specific modules, organised in pages, as further described next. The architecture also includes utility modules, such as a class for logging events and various plotting and metric utilities that provide cross-cutting support across the application.

## 2.2 User Interface and Additional Features

The graphical user interface (GUI) of Helix is implemented using Streamlit, a library designed for the development of data-centred applications. The decision to adopt Streamlit was due to its operating system agnosticism, lightweight deployment model, minimised development overhead, portability, and native compatibility with browser-based environments. By serving the interface via localhost, Streamlit enables access to Helix directly through the web browser without requiring external servers or very complex installation steps (see Figure 2). This ensures broad accessibility across operating systems (Windows, MacOS, Linux) and facilitates adoption.

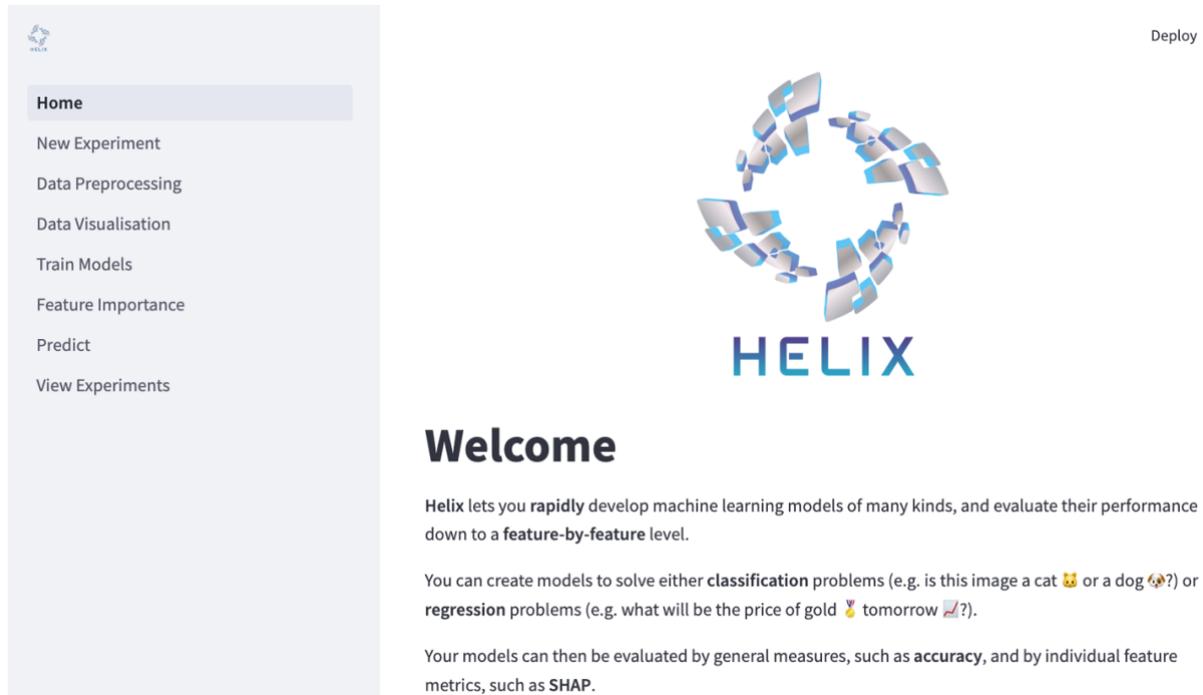

*Figure 2 - Screenshot of Helix's Welcome Page*

The interface in Helix is organised into modular components, namely pages, that align with the key stages of the analysis: experiment creation, data preprocessing, visualisation, model training, evaluation, ML interpretation and model deployment. Each module is rendered dynamically, allowing real-time interaction and immediate visual feedback.

## 2.2.1 Experiment Creation

In this module, a user defines general experiment parameters used throughout the pipeline. These include the name of the experiment, the data file, the name of the target variable (used for plotting and logging purposes), the problem type (either regression or classification) and a random seed to allow reproducibility of the results (Figure 3a).

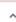

*Figure 3 - Screenshot of Helix's Create Experiment Page containing a) the experiment variables and b) the plot customisation options.*

This page also allows the user to set general settings to be applied to all plots generated (Figure 3b).

### 2.2.2 Data Preprocessing

*Data Preprocessing* is the first step of Helix's analytics pipeline. For the independent variables, normalisation methods currently available are standardisation (scale each independent variable between -1 and 1 values) and MinMax (scale each independent variable between 0 and 1).[9] Transformations for the dependent variable in regression problems include natural logarithm, square-root, standardisation and Minmax.

This page also allows the user to apply feature selection to their dataset.[10] The current methods supported are variance threshold (eliminate features that have a variance below a threshold given by the user), Pearson correlation threshold, where variables that have a correlation to another variable greater than a threshold given by the user are eliminated, and Least Absolute Shrinkage and Selection Operator (LASSO).

### 2.2.3 Data Visualisation

This module provides the user with a series of statistics and graphs for descriptive analytics. Both raw data and the processed data can be visualised. Examples of graphical descriptions are shown in Figure 4.

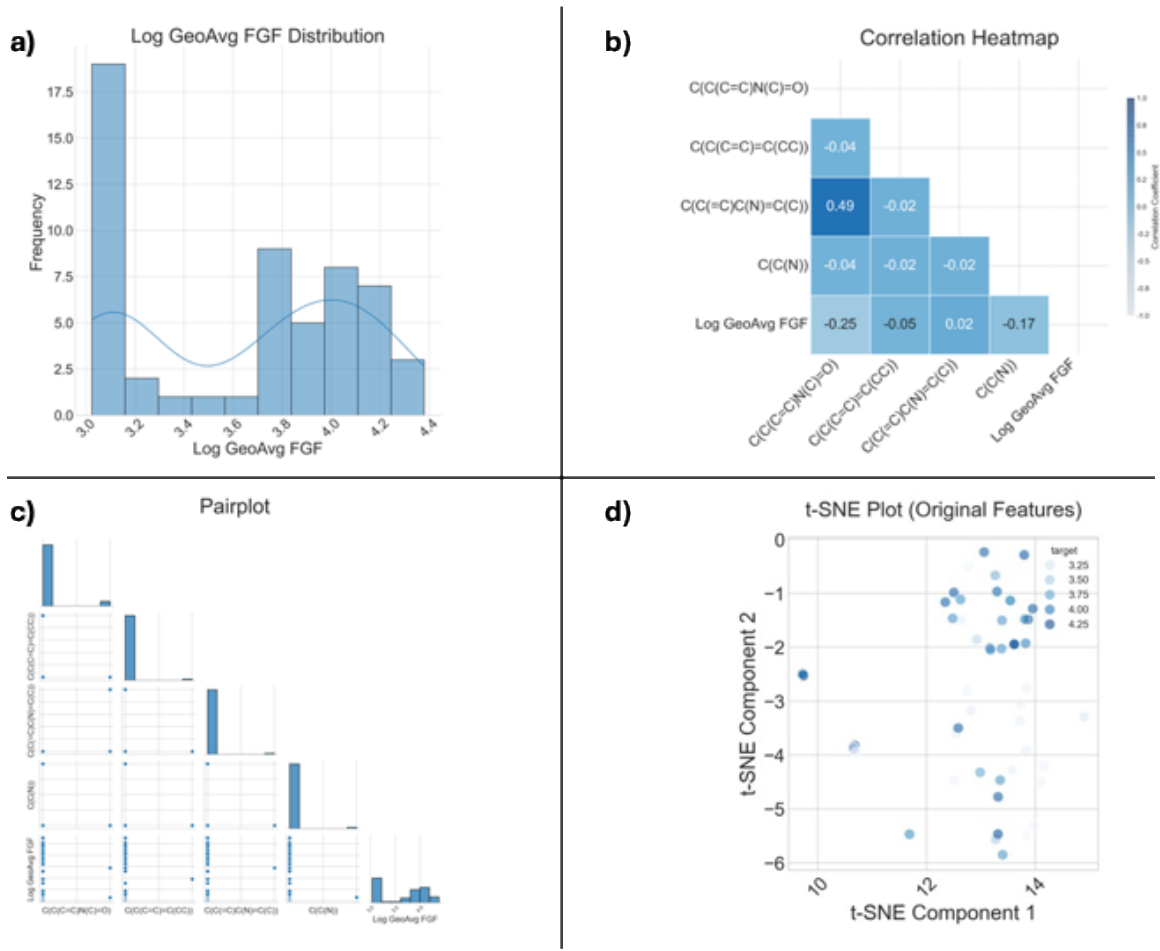

*Figure 4 - Examples of descriptions available in Helix. a) Distribution plot of the target variable, b) correlation heatmap, c) pairplot, and d) t-SNE plot.*

### 2.2.4 Machine Learning Modelling

In this page, the user has different options for modelling, including whether to apply hyperparameter optimisation, how the data splitting is carried out, and which machine learning algorithms are used for the training. Currently supported machine learning algorithms are Random Forest,[11] Gradient Boosting,[12] Support Vector Machine[13] and Logistic Regression[14] for classification problems. For regression problems, the same algorithms are available, but instead of logistic regression, multiple linear regression[15] and multiple linear regression with expectation maximisation[16] are available. To train the models, Helix gives the option to either perform hyperparameter optimisation using grid search or to set the hyperparameters manually (Figure 5).

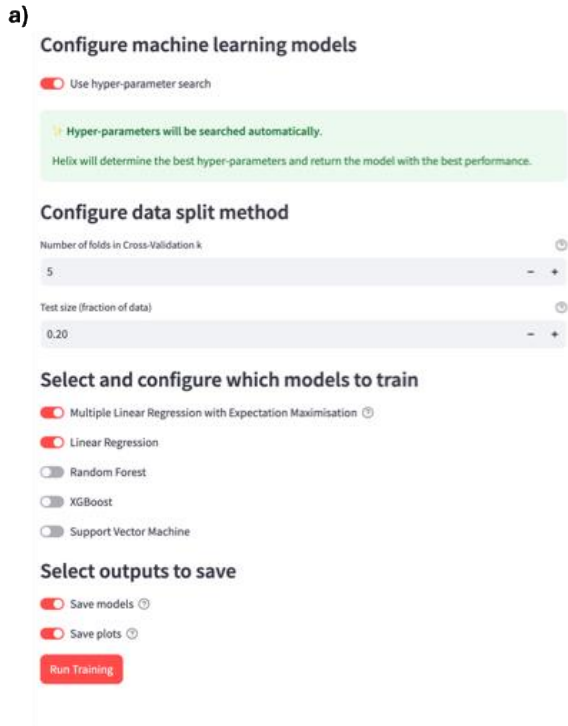
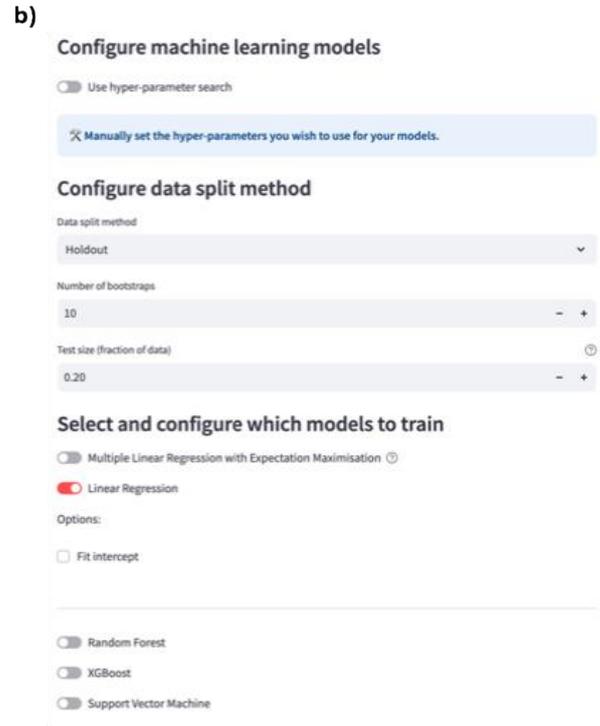

*Figure 5 - Machine learning model page options. a) Options available when running hyperparameter optimisation. b) Options shown to the user when parameters are selected manually for Linear Regression.*

Once the ML training and testing is finalised, results containing performance metrics for each algorithm are shown together with a table with the predictions of each of the models for each of the points (Figure 6a), and plots to show the performance of the models, as shown in Figure 6b.

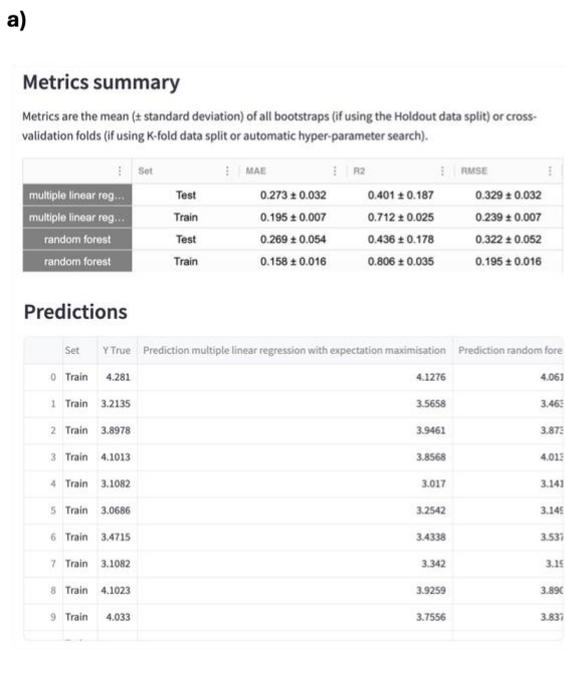
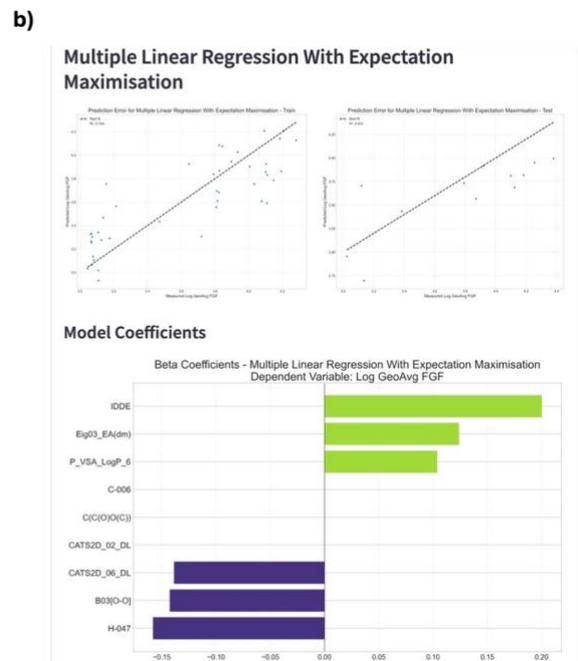

*Figure 6 - Example of Helix's output after machine learning model training and evaluation. a) Metrics obtained by the models and the instance-based predictions. b) Parity plot for each set and the beta coefficients for the linear models.*

## 2.2.5 Model Interpretation

Information such as what are the variables that a model considers most important when delivering a prediction or what is the impact of a variable's value to the final prediction are provided in this page. Algorithms implemented include global feature importance methods, such as permutation importance and SHAP and local feature importance methods, such as LIME and local SHAP. Helix also provides the user the option to get ensemble feature importance, which effectively takes the importance of each feature obtained from the global methods for all the models and fuses it into a single crisp importance score, which can be calculated for example using the mean importance value. Figure 7 shows examples of interpretation results visualisation in Helix.

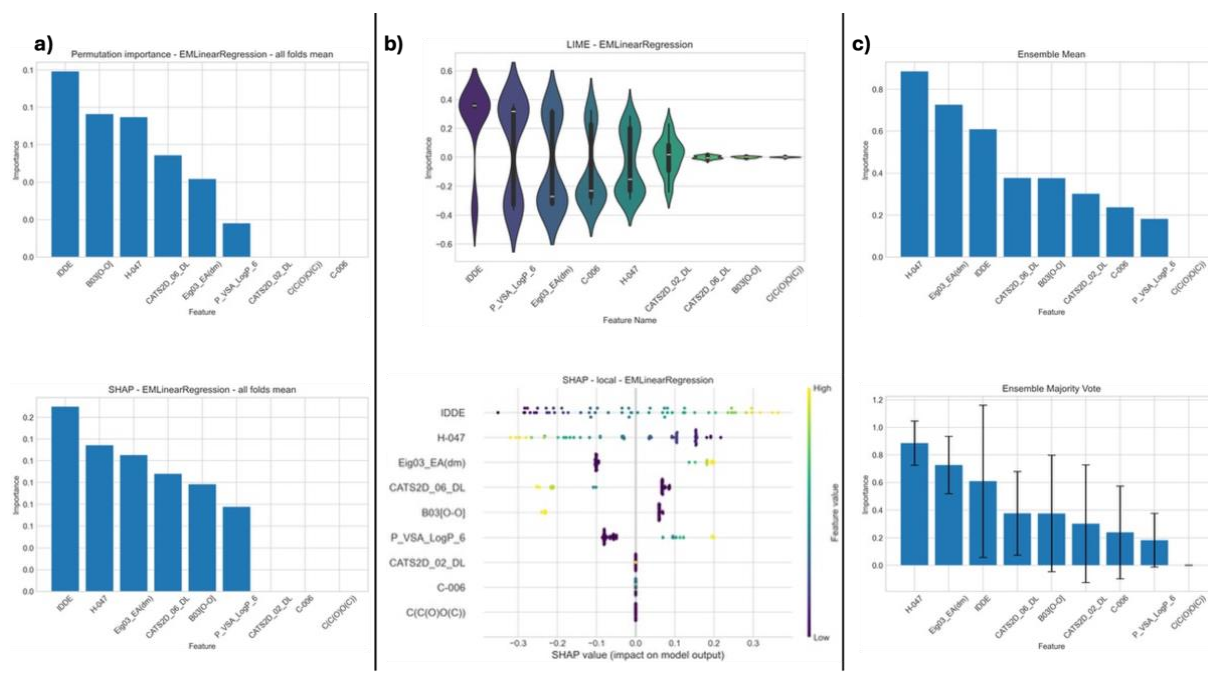

*Figure 7 - Interpretation results outputs. a) Plots for global feature importance, b) plots for local feature importance, and c) Ensemble feature importance plots.*

The fuzzy feature importance fusion uses a fuzzy logic-driven procedure adapted from[7] for interpreting and contextualising local feature importance. It selects key features based on user choices or their crisp importance ranks, transforms them into fuzzy categories, computes local importance scores, extracts fuzzy rules, and finally identifies recurring patterns linking these fuzzy sets to target outcomes and explaining them using natural language, if-then rules. This allows data and experimental scientists to understand feature synergy and importance in a more nuanced, human-readable manner, particularly when dealing with complex, non-linear models.

## 2.2.6 Model Deployment

The *Predict* page allows the user to use the trained models to predict the target variable on their data. For this, the user must provide a csv containing all the independent variables that they provided in the original training data. The user will be given the option to select which models are generating new predictions (see Figure 9a). For this, Helix applies the same transformations as those from the original data before and provide predictions for each model chosen.

## 2.2.7 Experiment Inspection and Automatic Track of Analysis Provenance

Helix offers a page where all analytics results can be visualised (Figure 8). This feature was created to facilitate team communication and analytics workflow auditing, as all results are summarised together will all parameters and options selected in previous pages.

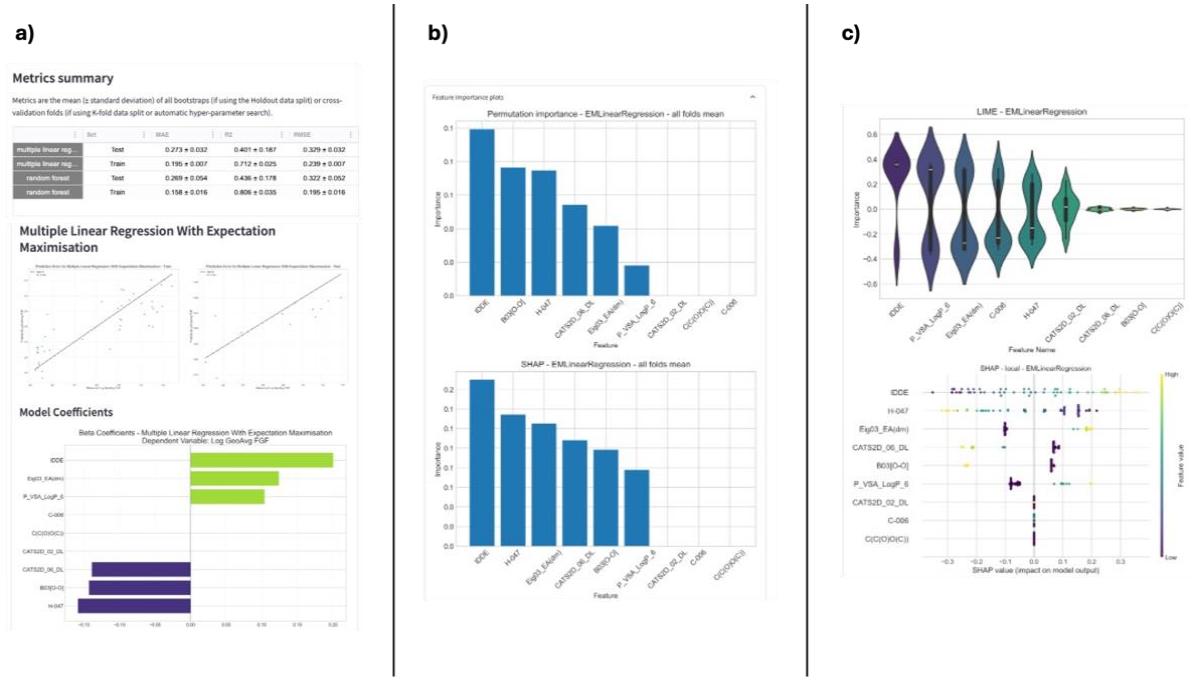

Figure 8 - Experiment visualisation outputs. a) Machine learning, b) global feature importance, and c) local feature importance results.

*Figure 9 - User preferences and parameters stored in Helix interface for all analytics stages. a) Execution and data options, b) plotting choices, c) preprocessing and machine learning selected and hyperparameters, and d) feature importance.*

For provenance and workflow metadata standardisation, the entire analytics experiment, including raw and pre-processed data, preprocessing choices, models, parameters, graphs, metrics are saved in a local folder and can be further inspected via user interface (as shown in Figure 8 and Figure 9) or by browsing a local directory file (Figure 10). The experiment folder can be shared and loaded in any machine that has Helix installed. Log files containing all events that occurred while running the tool, including user choices, analysis workflow steps taken and exceptions are also saved to support developers.

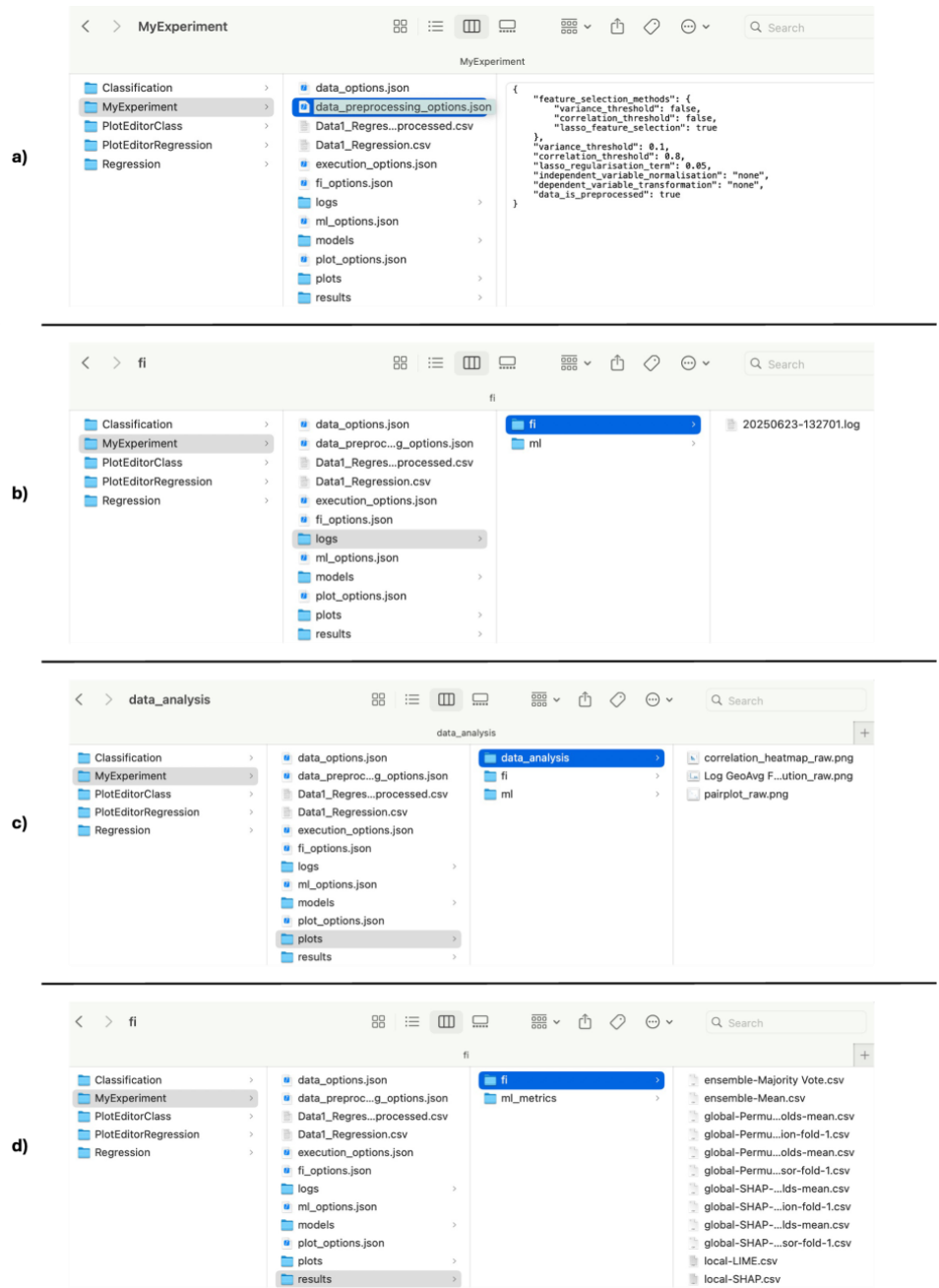

*Figure 10 - Experiment directory structure. a) Parent directory with all logs and results, b) example of a log file for machine learning experiments, c) example of plots for the data analysis subdirectory, and d) example of files structure for feature importance.*

## 2.3 Use Cases

### 2.3.1 Biomaterials

Results of analyses to identify microtopographical properties affecting biofilm formation using Helix have been recently published in[17]. In their work, Romero et al. investigated how 2176 topographical biomaterial shapes embossed into polymers reduced colonisation by two types of bacteria, P. *aeruginosa* and S. *aureus*. Using Helix to perform data processing and machine learning modelling and interpretation modules, the authors were able to successfully create predictive models and extract design rules influencing biofilm resistance for both cases.

### 2.3.2 Chemistry

To demonstrate Helix's capabilities in chemical sciences, we investigated the Delaney Solubility database.[18] This database comprises the solubility of 1144 organic compounds, expressed as the logarithm of the solubility in mols per litre. We used the molecule's minimum degree, molecular weight, number of hydrogen bond donors, number of rings, number of rotable bonds, and polar surface area. The dataset with the descriptors and the target variable has been sourced from the deepchem repository.[19] No preprocessing of the data has been performed. The results of modelling this dataset using linear regression with expectation maximisation are shown in **Figure 8**.

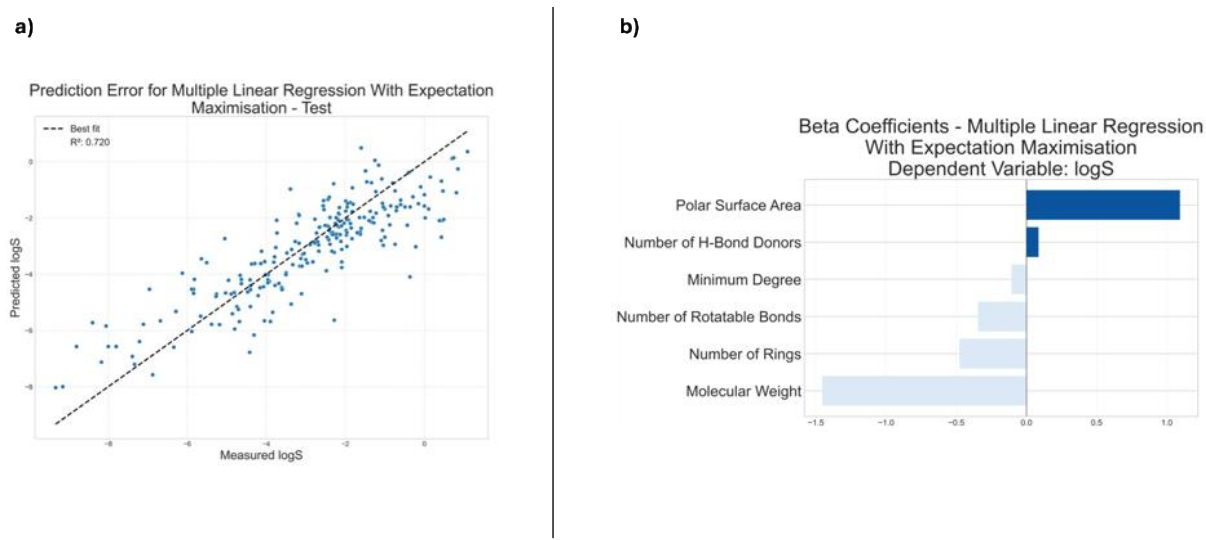

**Figure 11.** Results of modelling the Delaney Solubility dataset with Multiple Linear Regression with Expectation Maximisation using Helix. **a)** Parity plot of predicted and real logS values. **B)** coefficients of the multiple linear regression.

Helix delivered similar results to those reported previously (R2 of 0.720 in our case compared to 0.75 reported by Delaney).[18] By analysing the coefficients of the linear regression, it is possible to understand the impact of each of the variables for the solubility of the compounds. The polar surface area of the molecule and its number of hydrogen bond donors are the variables that the model found to contribute to a larger solubility, while the minimum degree, number of rotable bonds, number of rings and molecular weight decrease it. This makes sense from a chemical perspective, as the polar surface area and hydrogen bond donors are characteristics that make a molecule interact more strongly with molecules of water. On the other hand, properties such as molecular weight, number of rings and number of rotatable bonds imply that the molecule consists of big aliphatic groups that interact unfavourably with polar molecules, such as water.

### 2.3.3 Medicine

Helix was used in a real-world dataset provided by the Wellcome Leap in Utero SWIRL project.[20] This dataset includes clinical variables collected during pregnancy and aims to predict the risk of foetal demise (stillbirth), making it a high-stakes and sensitive classification task. It consists of 46 samples, each with 90 clinical features. It is highly imbalanced, containing 11 positive cases (classified as near miss) and 35 negative cases (healthy foetal outcomes). To mitigate the risk of overfitting due to the high number of variables relative to the sample size, Helix was used to conduct a two-stage modelling pipeline, with both stages incorporating min-max scaling to normalise feature values to the [0, 1] range.

**Stage 1: Feature Selection.** In the first stage, we employed all four classification models currently supported by Helix - Logistic Regression, Random Forest, XGBoost, and SVM - to perform 5-fold cross-validation on the complete dataset. Feature importance was then extracted using Helix's internal methods, and the results from all models were ensembled using majority voting to select the top five most important features (Figure 11).

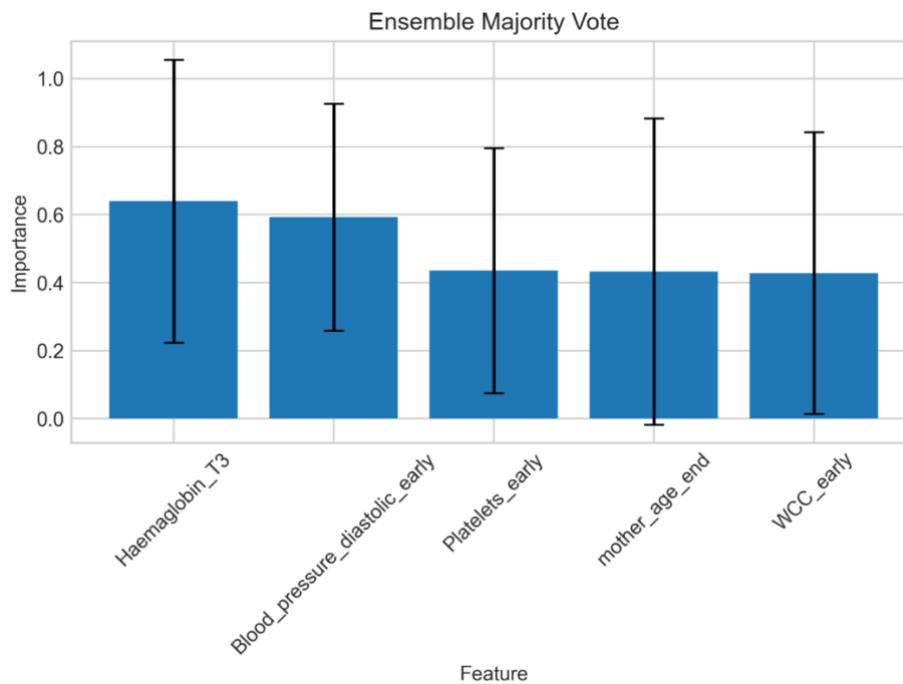

*Figure 11- Top-5 features identified from the SWIRL dataset based on a majority vote ensemble of feature importances from 5-fold cross-validation across four classifiers.*

The selected features and their clinical descriptions are summarised in Table 1:

*Table 1- Description of the top 5 selected features from the SWIRL dataset*

| Feature name | Description |
| --- | --- |
| Haemaglobin_T3 | Haemaglobin measurement in pregnancy at around 28 weeks gestation (grams/decilitre) |
| Blood_pressure_diastolic_early | First diastolic blood pressure measurement taken in pregnancy |
| Platelets_early | FIrst platelet count (per x 10^9/L) |
| mother_age_end | Mother's age at pregnancy end in complete years |
| WCC_early | FIrst white cell count (per x 10^9/L) |

**Stage 2: Classification.** In the second stage, a Logistic Regression model was trained using the top-5 selected features. We again applied 5-fold cross-validation on the full dataset. This yielded an accuracy of 0.815 ± 0.083 and an F1 score of 0.615 ± 0.509. The performance metrics and model interpretability are illustrated in Figure 12 - Results of the second-stage analysis on the SWIRL dataset using Logistic Regression with the top 5 features. (a) ROC Curve, (b)

SHAP plot illustrating the impact of each feature on near miss predictions., which includes both the ROC curve and the SHAP summary plot.

The SHAP plot reveals that higher diastolic blood pressure, platelet count, white cell counts in early pregnancy, and higher haemoglobin levels around 28 weeks are all positively associated with near miss cases. Interestingly, the model also suggests that younger maternal age may increase the likelihood of a near miss, a pattern that was later partially validated through expert review. However, experts also noted that this trend could be a result of sample bias due to the limited data size.

Overall, this case study highlights how Helix can aid experts in uncovering actionable patterns in clinical data, streamline feature selection, and reduce the complexity of medical data analysis. More importantly, Helix was able to expedite the analysis to inform further actions within the research team, including providing proof-of-concept validation of initial hypothesis around variable importance and how additional data collection should take place.

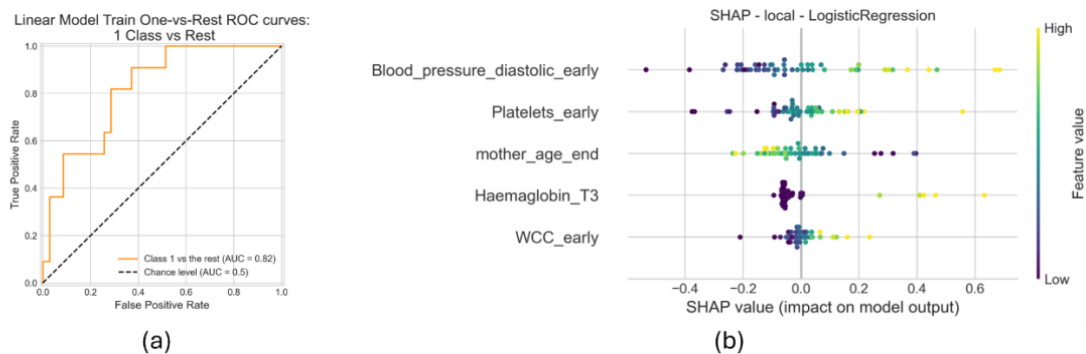

*Figure 12 - Results of the second-stage analysis on the SWIRL dataset using Logistic Regression with the top 5 features. (a) ROC Curve, (b) SHAP plot illustrating the impact of each feature on near miss predictions.*

## 3. FAIR Principles and Open Science

In alignment with the FAIR (Findable, Accessible, Interoperable, Reusable) principles,[21,22] Helix's source code is publicly available on GitHub, with comprehensive documentation hosted online. The software is distributed under the MIT licence, encouraging reuse and modification. Installation is facilitated via PyPI, and the platform supports integration with other Python-based tools, enhancing interoperability. Helix also adopts FAIR principles by systematically capturing and storing the provenance of data analytics workflows. By logging detailed metadata at each stage of the machine learning pipeline—including preprocessing decisions, model configurations, performance metrics, and feature importance outcomes—Helix ensures that analytical processes are transparent, traceable, and reproducible. This provenance information enables researchers to revisit and audit prior analyses and facilitates knowledge transfer and reuse across team members, projects, and institutions. The structured recording of analytical choices supports interoperability by enabling the integration of Helix outputs into broader computational ecosystems.

# 4. Conclusions

Helix represents a significant contribution to the suite of tools available for scientific data analysis, offering a reproducible and interpretable framework for ML on tabular data. It has been successfully applied to several domains, including biomaterials science, chemistry and medicine. Its open-source nature and adherence to FAIR principles, including a full record of analytics provenance adds value for researchers across multiple disciplines. Future developments aim to expand its capabilities to allow for more approaches to data pre-processing, modelling and interpretation. The use of Streamlit although makes Helix easy to use, presents limitations. While well-suited for lightweight, single-user applications, it currently lacks robust support for multi-user access control, concurrent sessions, and advanced state management—features that may be required for larger collaborative deployments or enterprise-scale systems.  Future development will aim at addressing these shortcomings and improve big data handling.

**Software Availability**

- **Source Code:** https://github.com/Biomaterials-for-Medical-Devices-AI/HelixGitHub+12GitHub+12GitHub+12
- **Documentation:** https://biomaterials-for-medical-devices-ai.github.io/Helix/index.html
- **PyPI Package:** https://pypi.org/project/helix-ai/PyPI

**Acknowledgements**

This work was supported by the Engineering and Physical Sciences Research Council (EPSRC) under the Materials for Medical Devices programme.